\documentclass[twocolumn, switch]{article} 

\usepackage{preprint}

\usepackage{amsmath, amsthm, amssymb, amsfonts}

\usepackage[numbers,square]{natbib}
\usepackage[utf8]{inputenc}	
\usepackage[T1]{fontenc}	
\usepackage{xcolor}		
\usepackage[colorlinks = true,
            linkcolor = purple,
            urlcolor  = blue,
            citecolor = cyan,
            anchorcolor = black]{hyperref}	
\usepackage{booktabs} 		
\usepackage{nicefrac}		
\usepackage{microtype}		
\usepackage{lineno}		
\usepackage{float}			

\usepackage{lipsum}		

\usepackage{newfloat}
\DeclareFloatingEnvironment[name={Supplementary Figure}]{suppfigure}
\usepackage{sidecap}
\sidecaptionvpos{figure}{c}

\usepackage{titlesec}
\titlespacing\section{0pt}{12pt plus 3pt minus 3pt}{1pt plus 1pt minus 1pt}
\titlespacing\subsection{0pt}{10pt plus 3pt minus 3pt}{1pt plus 1pt minus 1pt}
\titlespacing\subsubsection{0pt}{8pt plus 3pt minus 3pt}{1pt plus 1pt minus 1pt}

\usepackage{siunitx}
\usepackage{multirow}

\newcommand{\etal}{\textit{et al}.}
\newcommand{\ie}{\textit{i}.\textit{e}.}
\newcommand{\eg}{\textit{e}.\textit{g}.}

\newcommand{\myparagraph}[1]{\vspace{4pt}\noindent{\bf #1}}

\usepackage{xparse}
\ExplSyntaxOn
\NewDocumentCommand{\Rowvec}{ O{,} m }
 {
  \vector_main:nnnn { p } { & } { #1 } { #2 }
 }
\NewDocumentCommand{\Colvec}{ O{,} m }
 {
  \vector_main:nnnn { p } { \\ } { #1 } { #2 }
 }

\seq_new:N \l__vector_arg_seq
\cs_new_protected:Npn \vector_main:nnnn #1 #2 #3 #4
 {
  \seq_set_split:Nnn \l__vector_arg_seq { #3 } { #4 }
  \begin{#1matrix}
  \seq_use:Nnnn \l__vector_arg_seq { #2 } { #2 } { #2 }
  \end{#1matrix}
 }
\ExplSyntaxOff

\title{Periphery-Fovea Multi-Resolution Driving Model guided by Human Attention}

\usepackage{xwatermark}
\newwatermark[firstpage,color=gray!60,angle=90,scale=0.32, xpos=-4.05in,ypos=0]{\href{https://doi.org/}{\color{gray}{Publication doi}}}
\newwatermark[firstpage,color=gray!60,angle=90,scale=0.32, xpos=3.9in,ypos=0]{\href{https://doi.org/}{\color{gray}{Preprint doi}}}
\newwatermark[firstpage,color=gray!90,angle=0,scale=0.28, xpos=0in,ypos=-5in]{*correspondence: \texttt{yexia@berkeley.edu}}

\usepackage{authblk}

\author{Ye Xia}
\author{Jinkyu Kim}
\author{John Canny}
\author{Karl Zipser}
\author{David Whitney}

\affil{University of California, Berkeley}

\begin{document}

\twocolumn[ 
  \begin{@twocolumnfalse} 
  
\maketitle

\begin{abstract}
Inspired by human vision, we propose a new periphery-fovea multi-resolution driving model that predicts vehicle speed from dash camera videos. The peripheral vision module of the model processes the full video frames in low resolution. Its foveal vision module selects sub-regions and uses high-resolution input from those regions to improve its driving performance. We train the fovea selection module with supervision from driver gaze. We show that adding high-resolution input from predicted human driver gaze locations significantly improves the driving accuracy of the model. Our periphery-fovea multi-resolution model outperforms a uni-resolution periphery-only model that has the same amount of floating-point operations. More importantly, we demonstrate that our driving model achieves a significantly higher performance gain in pedestrian-involved critical situations than in other non-critical situations.
\end{abstract}
\vspace{0.35cm}

  \end{@twocolumnfalse} 
] 



\section{Introduction}
Vision-based deep autonomous driving models have shown promising results recently~\cite{kim2017interpretable, kim2018textual, bojarski2016end,xu2016end}. However, their performance is still far behind humans. An important aspect of human vision that distinguishes it from existing autonomous driving models is its multi-resolution property, with distinct foveal and peripheral structures that carry high-resolution and low-resolution information, respectively. The human fovea covers approximately two degrees of the central visual field; the rest of our visual field, \ie, the periphery, is blurry. Eye movements, guided by visual attention, are therefore necessary to gather high resolution foveal information from different parts of the visual field. One advantage of this design is its efficiency: resources are saved for particularly salient or important regions in what are otherwise redundant visual scenes. Driving scenes seem to be highly redundant, as well, considering the large portions of uniform areas such as the sky, buildings, and roads. Inspired by the human vision, we propose a new periphery-fovea multi-resolution driving model and show that it achieves higher driving accuracy and better efficiency.

The first challenge in designing this model is to effectively combine the global low-resolution peripheral vision and the local high-resolution foveal vision that dynamically scans across the frame. We propose two ways to merge the two visions by either using a combined peripheral-foveal planner or two independent visual planners. We will compare their performances and discuss the differences.

\begin{figure}[!t]
    \begin{center}
        \includegraphics[width=\linewidth]{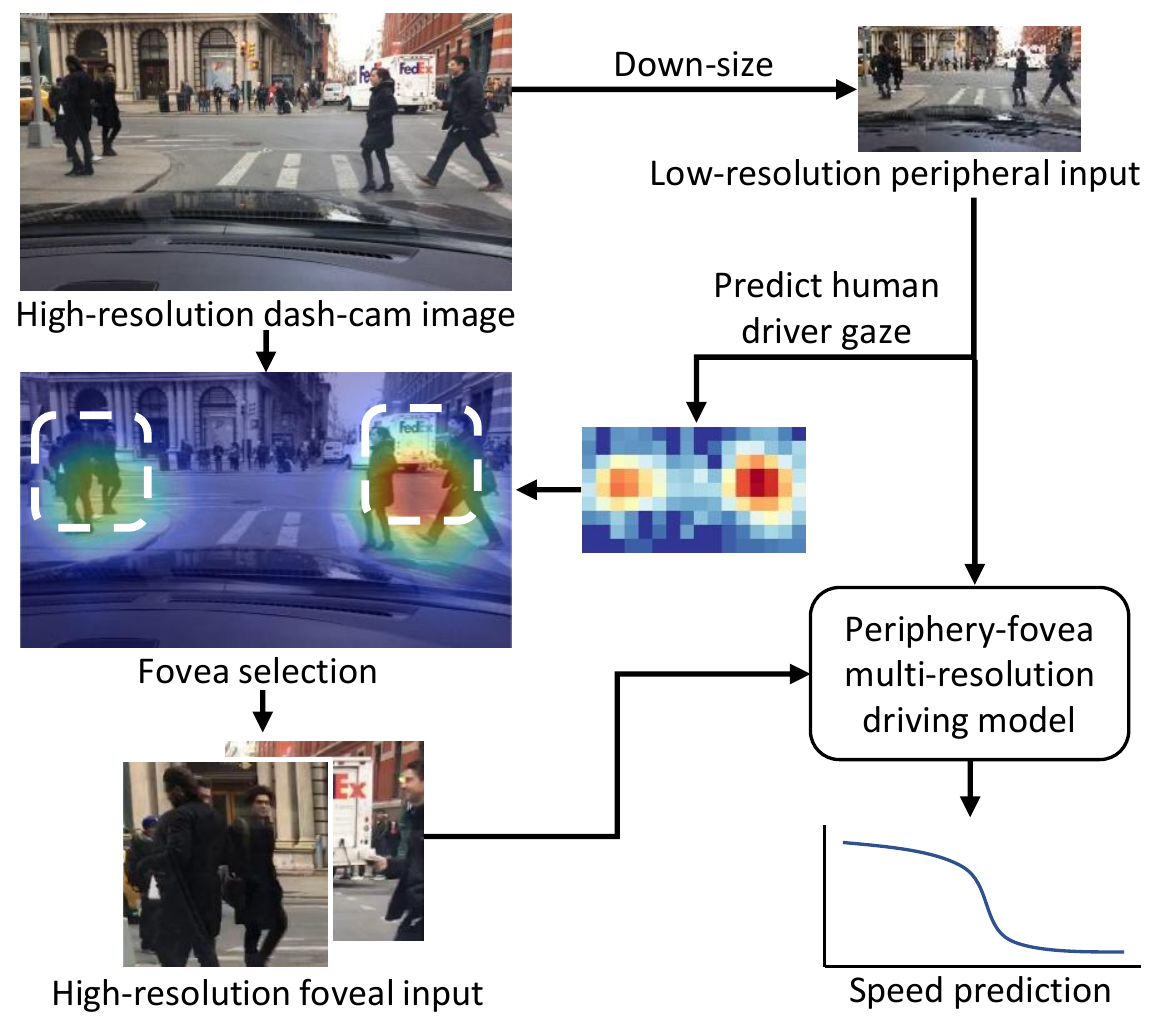}
    \end{center}
    \caption{Our model uses the low-resolution full video frame as the peripheral visual input to predict human driver gaze and gets high-resolution image patches from the predicted gaze locations. It then combines the peripheral input and foveal input to predict the vehicle speed at high accuracy and high efficiency.}
    \label{fig:teaser} 
\end{figure}

The second challenge is how to dynamically guide foveal vision to the critical locations. The foveal location selection is a non-differentiable process. A potential solution is to use reinforcement learning, but it could take a great deal of data and training. We choose a different approach: guiding the foveal vision to where human drivers would gaze. Recently proposed large driver gaze datasets~\cite{xia2017predicting,dreyeve2016} and driver gaze prediction models~\cite{xia2017predicting,palazzi2018predicting,palazzi2017learning} allow us to predict human gaze for our videos. However, it has not been tested whether predicted human gaze or even ground-truth human gaze can benefit autonomous driving models. Note that in order to be highly efficient, the human gaze can only be predicted using low-resolution input images, which makes the question even more complex.

A unique property of human gaze is that it reveals the relative urgency of locations and objects of potential interest. Different moments during driving and different road agents are not equally urgent. Human drivers look at the most critical regions when emergencies arise. Incorporating human gaze into a driving model may not only increase its average performance but also bring even higher performance gain at critical moments. We use a driving video dataset that has human-annotated explanations about the driver's actions. We demonstrate that our driving model guided by human gaze shows even higher performance gain in the cases where reactions to pedestrians are necessary than in other presumably less critical cases.

\section{Related work}
\myparagraph{End-to-End Learning for Self-driving Vehicles.}
Recent successes~\cite{bojarski2016end,xu2016end} suggest that a driving policy can be successfully learned by neural networks with the supervision of observation (\ie raw images)-action (\ie steering) pairs collected from human demonstration. Bojarski~\etal~\cite{bojarski2016end} trained a deep neural network to map a dashcam image to steering controls, while Xu~\etal~\cite{xu2016end} utilized a dilated deep neural network to predict a vehicle's discretized future motions. Hecker~\etal~\cite{hecker2018end} explored an end-to-end driving model that consists of a surround-view multi-camera system, a route planner, and a CAN bus reader. Explainability of deep neural networks has been increasingly explored. Kim~\etal~\cite{kim2017interpretable, kim2018textual} explored an interpretable end-to-end driving model that explains the rationale behind the vehicle controller by visualizing attention heat maps and generating textual explanation. Recently, Wang~\etal~\cite{wang2018deep} introduced an instance-level attention model that finds objects (\ie, cars, and pedestrians) that the network needs to pay attention to.

\myparagraph{Incorporating human visual attention.}
Attention mechanisms have shown promising results in various computer vision tasks, \eg, image caption generation~\cite{xu2015show}, visual question answering (VQA)~\cite{zhu2016visual7w}, and image generation~\cite{gregor2015draw}. Most of these models do not supervise the generated attention by human attention. Recently, Das~\etal~\cite{das2017human} has shown that explicitly supervising the attention of VQA models by human attention improves the models' VQA performance. Zhang~\etal~\cite{zhang2018agil} has trained a network that predicts human attention for Altari games and shown that incorporating the predicted human attention into the policy network significantly improves the action prediction accuracy. However, incorporating human visual attention in driving tasks has not yet been explored. Besides, the previously mentioned attention models use high-resolution images to generate attention. Predicting attention using low-resolution input and combining global low-resolution input and attended local high-resolution input has not been explored.

\myparagraph{Predicting driver attention.}
Recently, deep driver attention prediction models~\cite{xia2017predicting,palazzi2018predicting,palazzi2017learning} have been proposed. The input of these models is video recorded by cameras mounted on the car. The output is an attention map indicating the driver's gaze probability distribution over the camera frame. These models are trained using large-scale driver attention datasets~\cite{xia2017predicting,dreyeve2016} collected with eye trackers, and they use high-resolution input images ($576\times1024$ or higher) to achieve optimal accuracy. How reliable the prediction would be using low-resolution input images have not been explored.

\begin{figure*}
\begin{center}
\includegraphics[width=\linewidth]{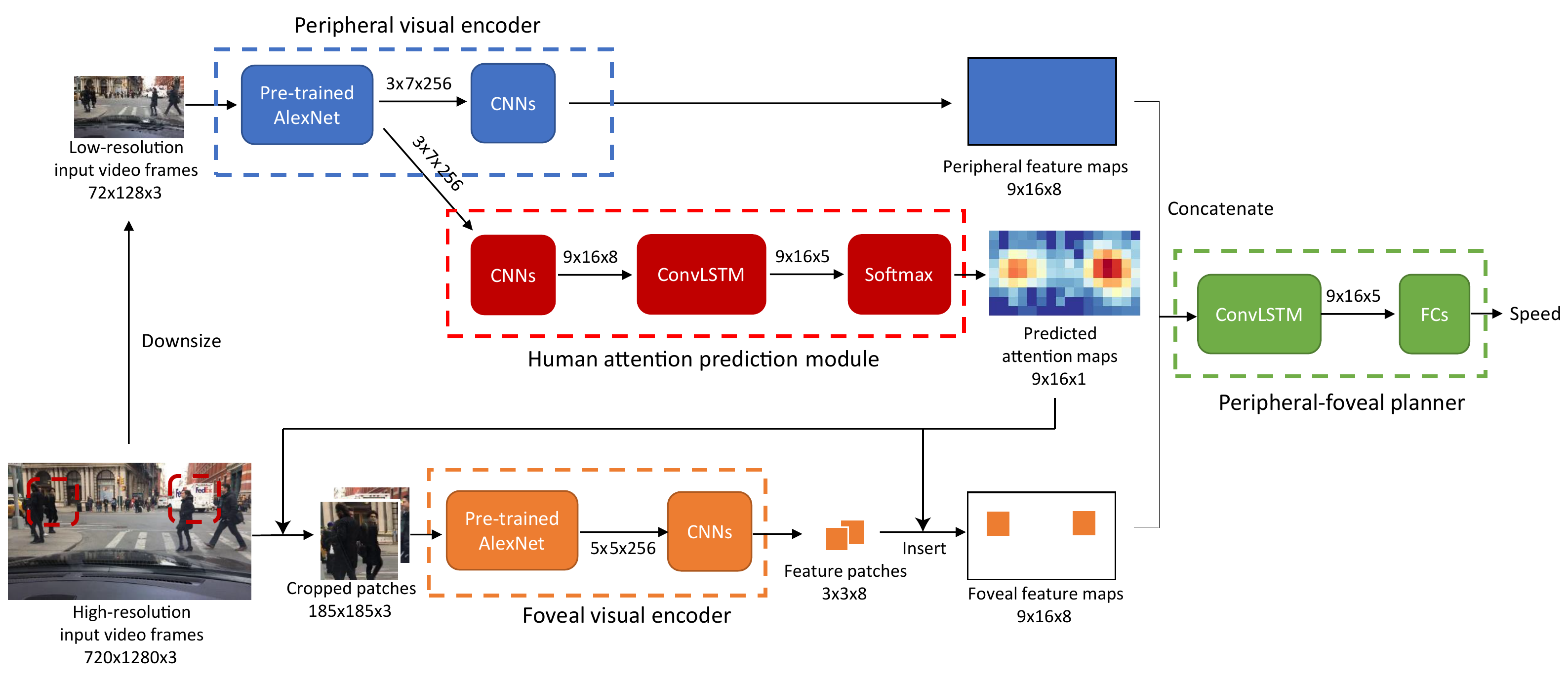} 
\end{center}
   \caption{Our model consists of four parts: (1) the peripheral visual encoder, which extracts high-level convolutional visual features (CNN here); (2) the human attention prediction module, which learns the behavior of human attention as a supervised learner over image-gaze pairs collected from humans; (3) the foveal visual encoder, which selects fovea locations, crops the high-resolution fovea image patches and encodes them into visual features; (4) the peripheral-foveal planner, which combines the peripheral and foveal visual features and predicts a low-level control command, \ie, a vehicle's speed.}
\label{fig:model}
\end{figure*}

\section{Periphery-Foveal Multi-Resolution Model}
Here, we propose a novel driving model that mimics the key aspect of the human vision system: the peripheral and the foveal systems. Our model mainly uses the peripheral vision to predict a control command (\ie, speed) in an end-to-end manner, but we add the foveal vision to improve the model's perceptual primitives.  While the peripheral vision sees the whole but blurry image, the foveal vision fixates on parts of the images with a higher resolution. To this end, our model needs three main capabilities: (1) the ability to extract perceptual primitives to manipulate the vehicle's behavior, (2) the ability to find out image regions where the model needs to attend with a high resolution (\ie, pedestrians, traffic lights, construction cones, etc), (3) the ability to augment the peripheral vision system with the foveal vision. 

As we summarized in Figure~\ref{fig:model}, our model consists of four parts: (1) the {\em{peripheral visual encoder}}, which extracts high-level convolutional visual features (CNN here); (2) the {\em{human attention prediction module}}, which learns the behavior of human attention as a supervised learner over image-gaze pairs collected from humans; (3) the {\em{foveal visual encoder}}, which selects fovea locations, crops the high-resolution fovea image patches and extracts visual features from the high-resolution image patches; (4) the {\em{peripheral-foveal planner}}, which combines the peripheral and foveal visual features and predicts a low-level control command, \ie a vehicle's speed.

\subsection{Peripheral Visual Encoder}
We sample the video frames at 10 Hz. The original frame images have a resolution of $720\times1280$ pixels. We downsample them to $72\times128$ pixels as the input for the peripheral vision input of our model. The raw pixel values are subtracted by [123.68, 116.79, 103.939] as \cite{krizhevsky2012imagenet}.

The low-resolution frame images are first passed to the peripheral feature encoder. This feature encoder consists of an ImageNet pre-trained AlexNet and three additional convolutional layers. The weights of the pre-trained AlexNet are fixed and not further trained during the training of our driving model. Each of the additional convolutional layers is followed by Batch Normalization and Dropout. The output feature maps of this feature encoder have a size of $3\times7$ pixels and 8 channels. These feature maps are then upsampled to $9\times16$ pixels for the next steps.

\subsection{Human Attention Prediction Module}
The low-resolution frame images are also passed to a human attention prediction module to determine where human drivers would gaze. We used the model described in \cite{xia2017predicting} as our human attention prediction module. This model consists of a fixed ImageNet pre-trained AlexNet, three additional convolutional layers, and a Convolutional Long Short-Term Memory (ConvLSTM) module. Since both the peripheral feature encoder and the human attention prediction module start with passing the low-resolution through the same fixed AlexNet, this passway is shared by both modules. The human attention prediction module is separately trained using a human driver attention dataset and is fixed during the training of the driving model. The predicted human attention maps have a resolution of $9\times16$ pixels.

\subsection{Foveal Visual Encoder}
The foveal visual encoder chooses two independent fovea locations for each input frame. In the following experiments, the fovea locations can be chosen in four different ways: random selection over the frame, always selected from the frame center, a top-k method and a sampling method. The top-k method selects the two pixels that have the highest attention intensities in each predicted $9\times16$-pixel human attention map. The sampling method samples two fovea locations following the predicted attention probability distribution modulated by a temperature factor described by the following formula:
\begin{equation}
 p_{i} = \frac{ \exp{(\log q_{i}/T)} }{ \sum_{j}{\exp{(\log q_{j}/T)}} } 
\end{equation}
where $p_{i}$ is the probability of the i-th pixel being selected as the fovea location, $q_{i}$ is the predicted human attention probability at the i-th pixel, and $T$ is the temperature factor. A temperature factor of 1 means sampling faithfully following the predicted human attention distribution. A higher temperature factor means sampling more uniformly. A lower temperature factor means sampling more from the pixel that has the highest human attention intensity.

An image patch of $240\times240$ pixels centered at each selected fovea location is cropped out from the $720\times1280$-pixel high-resolution frame image. The images patches are then downsized to $185\times185$ pixels to fit the receptive fields and strides of the following encoder network. The raw pixel values are subtracted by [123.68, 116.79, 103.939] as \cite{krizhevsky2012imagenet} before being passed to the encoder network. The foveal visual encoder has the same structure as the peripheral visual encoder except for the kernel sizes and strides of the additional convolutional layers.

\subsection{Peripheral-Foveal Planner}
The peripheral-foveal planner further processes the peripheral and foveal features to predict speed for the future. It first creates a foveal feature map that has the same size as the peripheral feature map ($9\times16$ pixels, eight semantic channels). The foveal feature map is initialized with zeros. Each foveal image patch is encoded into a $3\times3\times8$ feature patch by the foveal feature encoder. These foveal feature patches ($\mathbf{y}_{i,j}$) are inserted into the foveal feature map ($\mathbf{x}_{i,j}^f$) at locations corresponding to the foveal locations: 
\begin{equation}
 \mathbf{x}_{i+h,j+w}^f = \mathbf{y}_{i,j}
\end{equation}
where $h$ and $w$ are the height and width coordinates of the top-left corner of the fovea patch.

In the cases where the feature patches of two foveae overlap, the maximum of each pair of overlapping feature values is kept. Then the peripheral feature maps ($\mathbf{x}_{i,j}^p$) and foveal feature maps ($\mathbf{x}_{i,j}^f$) are concatenated along the semantic dimension to form the combined feature maps ($\mathbf{x}_{i,j}^c$). 
\begin{equation}
 \mathbf{x}_{i,j}^c = \Colvec{\mathbf{x}_{i,j}^p,\mathbf{x}_{i,j}^f}
\end{equation}
The combined feature maps are then processed by a ConvLSTM layer and four fully-connected layers to predict a continuous value for the vehicle speed.

\section{Experiments}
In this section, we first present the datasets we used and our training and evaluation details. Then, we make quantitative and qualitative analyses of our proposed periphery-fovea multi-resolution driving model. 

\begin{figure*}[t]
\begin{center}
\includegraphics[width=.95\linewidth]{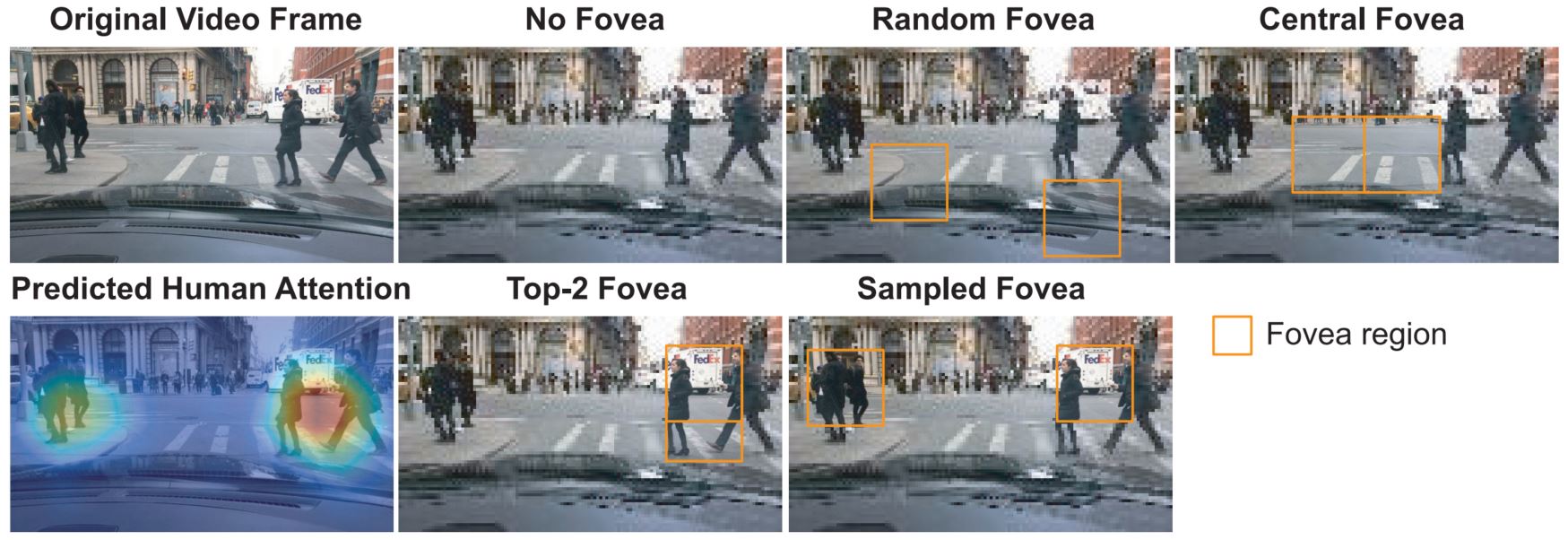} 
\end{center}
   \caption{
   Examples of different approaches of foveal region selection. We present the original input video frame and the predicted human attention heat map at the left column. Our baseline model only uses peripheral vision (without fovea). We studied four different types of foveal vision selection: random, central, top-2, and sampling. Top-2 and sampled foveae are chosen according to the predicted human attention. For better visualization, we present orange boxes to indicate the foveal regions.}
\label{fig:fovea}
\end{figure*}

\subsection{Datasets}
We used the Berkeley DeepDrive eXplanation (BDD-X) dataset~\cite{kim2018textual} to train and evaluate the driving models. This dataset contains human-demonstrated dashboard videos of urban driving scenes in various weather and lighting conditions. The dataset also provides a set of time-stamped sensor measurements, \eg, vehicle’s velocity and course, and time-stamped human annotations for vehicle action descriptions and justifications. The training set contains 5,588 videos and the validation and testing sets contain 698 videos. Most videos are 40 seconds long.

We used the Berkeley DeepDrive Attention (BDD-A) dataset~\cite{xia2017predicting} to train the human attention prediction module. The BDD-A dataset contains driving videos collected in the same way as the BDD-X dataset. (But the two datasets do not share the same videos.) The BDD-A dataset also provides human attention map annotations. The human attention maps were collected by averaging multiple drivers' eye movements while they were watching the videos and performing a driver instructor task~\cite{xia2017predicting}. The attention maps highlight where human drivers need to gaze when making driving decisions in the particular situations. The BDD-A dataset contains 926, 200 and 303 videos in the training, validation and testing sets, respectively. Each video is approximately 10-second-long.

\subsection{Training and Evaluation Details}
The AlexNet modules in the driving models were pre-trained on ImageNet and frozen afterwards. The human attention prediction module was trained following \cite{xia2017predicting} except that the input image resolution was $72\times128$ pixels. Other parts of the driving models were trained end-to-end from scratch. We used the Adam optimization algorithm~\cite{kingma2014adam}, dropout~\cite{srivastava2014dropout} at a drop rate of 0.2, and the Xavier initialization~\cite{glorot2010understanding}. The training of our model took approximately one day on one NVIDIA GeForce GTX 1080 GPU. Our implementation is based on Tensorflow \cite{abadi2016tensorflow} and our code will be publicly available upon publication. The models were set to predict the vehicle speed one second in the future. We used three metrics, \ie, the mean absolute error (MAE), the root-mean-square error (RMSE), and the correlation coefficient (Corr), to compare the prediction against the ground-truth speed signals to evaluate the performances of the driving models. At inference time, the longest single video duration that our GPU memory could process was 30 seconds. Therefore, during training, unless otherwise stated, the original testing videos that were longer than 30 seconds were divided into 30-second-long segments and the remaining segments.

\subsection{Effect of the foveal vision guided by human attention}
To test the effect of the foveal vision guided by human attention, we compared our peripheral-foveal multi-resolution driving model against three baseline models (Figure~\ref{fig:fovea}). The first baseline model (no fovea) uses only low-resolution full video frames as input and has only the peripheral branch of the driving model we introduced. The second baseline model (random fovea) select fovea locations randomly over the video frame. The third baseline model (central fovea) always assigns its two foveae to the central $240\times480$ region of the frame. The central-fovea model is a strong baseline because the central regions mostly cover the area the vehicle is driving into and human drivers mostly localize their attention around the center of the road. We compared these baseline models with our peripheral-foveal multi-resolution driving model guided by human attention (human-guided fovea). The fovea locations were selected using the top-2 method. The mean testing errors of these models are summarized in Table~\ref{tab:metrics}. Our driving model outperformed all of the baseline models. This result suggests that the foveal vision guided by predicted human attention can effectively improve the model's accuracy. Note that the random-fovea model performed worse than the no-fovea model. This suggests that adding high-resolution foveal input would not necessarily improve the model. If fovea locations are not selected in a proper way, it may add distracting information to the driving model.

{
\setlength{\tabcolsep}{4pt}
\renewcommand{\arraystretch}{1.3} 
\begin{table}[t]
	\begin{center}
	\caption{
	We compared the vehicle control (\ie speed) prediction performance of four different types of vision systems.  We evaluated their performance in terms of the mean absolute error (MAE), the root-mean-square error (RMSE), and the correlation coefficient (Corr).}\vspace{-1em}
     \label{tab:metrics}
    	\resizebox{\linewidth}{!}{%
    	\begin{tabular}{@{}lcccc@{}} \toprule
    	    \multirow{2}{*}{Model} & \multicolumn{3}{c}{Speed (km/h)} \\ \cmidrule{2-4}
             & MAE & RMSE & Corr \\ \midrule
            Peripheral vision only (no fovea, baseline) & 9.6 & 14.4 & .594\\
            w/ Random fovea & 11.2 & 15.4 & .520\\
            w/ Central fovea & 9.4 & 13.9 & .592\\
            w/ Human-guided fovea (ours) & \textbf{9.1} & \textbf{13.4} & \textbf{.596}\\
            \bottomrule
        \end{tabular}}
     \end{center}
\end{table}
}


\subsection{Sampling according to multi-focus human attention}
Human attention can be multi-focus~\cite{cavanagh2005tracking}, especially during driving when the driver needs to react to multiple road agents or objects. A concern about using the top-2 method to select fovea locations is that it may select adjacent locations around a single focus in one frame and also select locations from the same focus in the next frames. To address this concern, we brought a sampling method to select fovea locations (described in the Model section). It samples fovea locations according to the predicted human attention probability distribution and modulated by a temperature factor (Figure~\ref{fig:fovea}). We tested our driving model using both the top-2 method and the sampling method and experimented with three different temperature factor values for the sampling method. To quantify to how much extend the fovea selection followed the predicted human attention, we calculated the likelihood of the selected foveae. To quantify the redundancy in fovea location selection, we calculated the overlap ratio between the fovea patches of adjacent frames. The results are summarized in Table~\ref{tab:sampling}. The results showed the trend that a balance between high likelihood and low overlap would result in the optimal performance. In our experiments, sampling completely following the predicted human attention distribution (\ie, temperature factor $T = 1$) showed the best prediction accuracy. 

{
\begin{table}
	\begin{center}
	\caption{Mean testing errors of our driving model using different fovea selection methods.}\vspace{-1em}
     \label{tab:sampling}
    	\resizebox{1\linewidth}{!}{%
    	\begin{tabular}{@{}lcccccc@{}} \toprule
            Fovea selection & Temperature & Likelihood & Overlap & MAE & RMSE & Corr \\ \midrule
            Top-2 fovea   & -   & 0.48 &  92\% & 9.1 & 13.4 & .596\\
            Sampled fovea & 0.5 & 0.46 &  55\% & 8.6 & 12.7 & .622\\
            Sampled fovea & 1   & 0.37 &  32\% & \textbf{8.5} & \textbf{12.4} & \textbf{.626}\\
            Sampled fovea & 2   & 0.18 &  11\% & 8.7 & 12.9 & .621\\
            \bottomrule
        \end{tabular}}
     \end{center}
\end{table}
}

{
\begin{table}[t]
	\begin{center}
	\caption{Mean testing errors of our driving models using either combined or dual peripheral-foveal planner.}\vspace{-1em}
     \label{tab:combined_dual}
    	\resizebox{\linewidth}{!}{%
    	\begin{tabular}{@{}lcccc@{}} \toprule
            Model & MAE & RMSE & Corr \\ \midrule
            Ours w/ Dual Peripheral-foveal Planner& 9.4 & 13.2 & .602\\
            Ours w/ Combined Peripheral-foveal Planner& \textbf{8.5} & \textbf{12.4} & \textbf{.626}\\
            \bottomrule
        \end{tabular}}
     \end{center}
\end{table}
}

\subsection{Comparison between combined and dual peripheral-foveal planner}
The previously presented design of our peripheral-foveal planner combines peripheral and foveal features to process with one ConvLSTM network. We call this design the combined peripheral-foveal planner design. In this design, the peripheral and foveal feature maps need to have the same resolution in order to be concatenated along the semantic dimension ($9\times16$ in our case). This constraint determines that the feature patch corresponding to one foveal input image patch cannot be bigger than $3\times3$ pixels. 

To break this constraint, we experimented with a different design, \ie, the dual peripheral-foveal planner structure. It bypasses the uni-resolution constraint by processing the peripheral and foveal features with separate ConvLSTM networks. It generates a feature patch of $14\times14$ pixels for each foveal input image patch. In stead of inserting the foveal feature patch into a bigger grid that corresponded to the full video frame, it adds the positional encoding \cite{vaswani2017attention} of the fovea location into the fovea features to preserve the fovea location information.

We tested the dual planner and compared it against the combined planner. The dual planner did not show higher accuracy than the combined planner (Table~\ref{tab:combined_dual}). We think this is because the combined planner also have its own unique advantages. In the combined planner design, the fovea location is clearly indicated by the location of the features in the feature map. Besides, the foveal features and peripheral features that are calculated from the same frame region are aligned into one vector in the combined feature maps. So the kernel of the upcoming ConvLSTM network can process the peripheral and foveal features of the same region jointly.

\begin{figure}[t]
\begin{center}
\includegraphics[width=.8\linewidth]{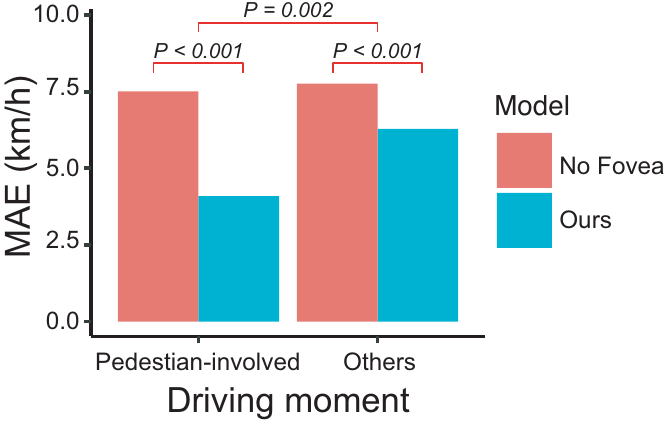}
\end{center}
   \caption{Testing errors of the no-fovea baseline model and our model at pedestrian-involved moments and other moments when the vehicle speed is under \SI{10}{\meter/\second} (\SI{36}{\km/\hour}). Statistical significance levels given by permutation tests are noted in the graph.}
\label{fig:pedestrian_error}
\end{figure}

\begin{figure*}[t]
\begin{center}
\includegraphics[width=.95\linewidth]{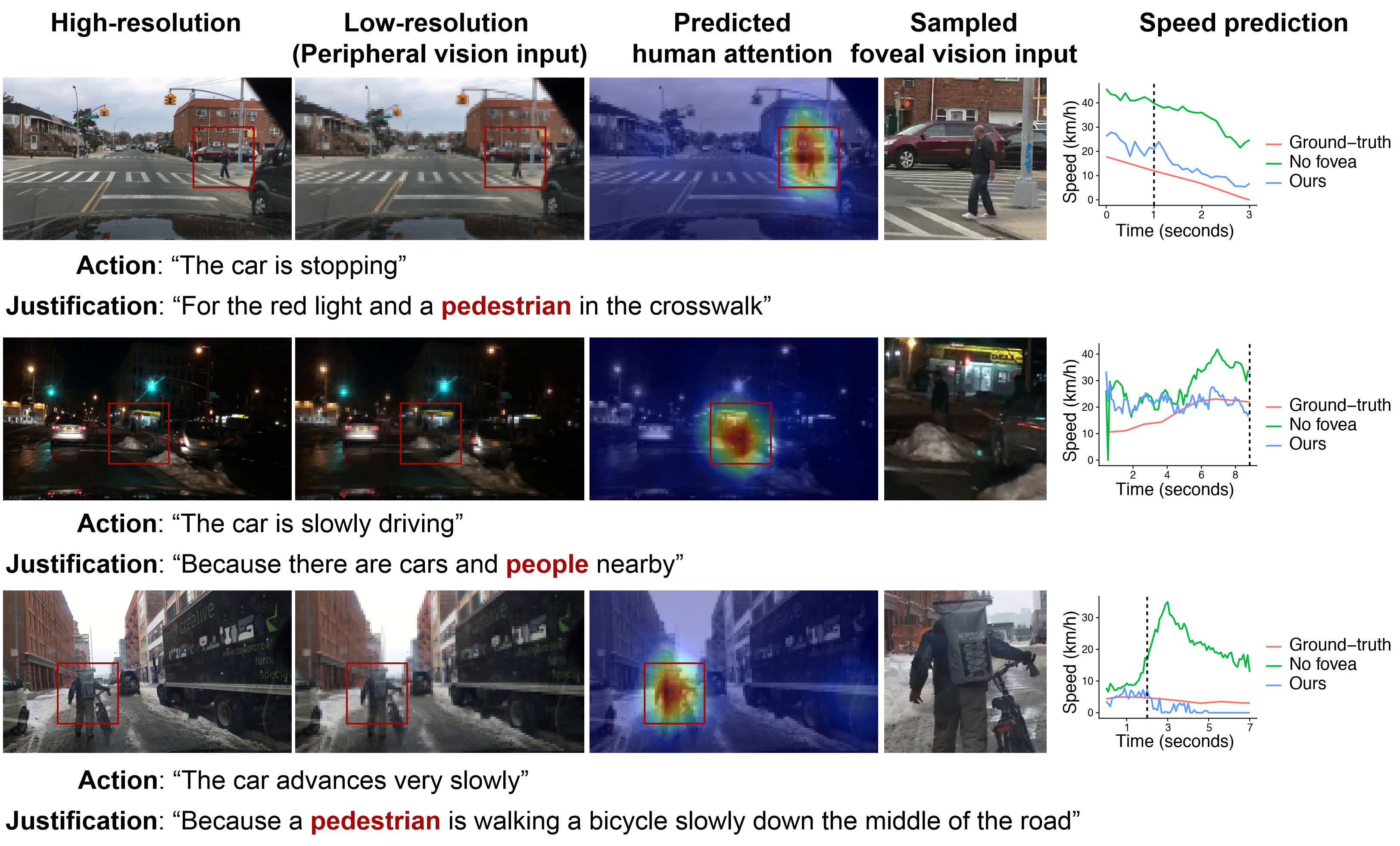} 
\end{center}
   \caption{Examples showing how our model and the no-fovea model react in pedestrian-involved situations. From left to right: original high-resolution frame images, low-resolution frame images used as peripheral vision input, predicted human attention maps, selected high-resolution image patches as foveal vision input, and ground-truth and predicted speed curves. The vertical dashed lines in the speed curve graphs indicate the moments depicted by the frame images. The textual action and justification human annotations are displayed below the images of each example.}
\label{fig:examples}
\end{figure*}

\subsection{Larger performance gain in pedestrian-involved critical situations}

The textual annotations of the BDD-X dataset allowed us to identify the critical situations where the driver had to react to pedestrians. These pedestrian-involved situations were defined as the video segments where the justification annotations contained the word "pedestrian", "person" or "people". We tested whether our model showed a stronger performance gain in the pedestrian-involved situations than in the remaining situations which should be on average less critical.

We calculated the mean prediction errors of our model and the no-fovea model separately for the pedestrian-involved video segments and the remaining segments in the test set. Note that the prediction error correlates with the vehicle speed and the pedestrian-involved segments only covered a speed range up to \SI{10}{\meter/\second} (\SI{36}{\km/\hour}). For a fair comparison, we excluded the frames in which the vehicle speed was higher than \SI{10}{\meter/\second} from this analysis. In order to determine the statistical significance levels, we ran permutation tests that could address the concern that the frames of a video are not independent.

The results are summarized in Figure~\ref{fig:pedestrian_error}. Our model showed significant performance gains in both the pedestrian-involved situations and the remaining situations (P value $< 0.001$). More importantly, the gain achieved in the pedestrian-involved situations was significantly bigger than the gain in the remaining situations (P value $= 0.002$). Some examples are demonstrated in Figure~\ref{fig:examples}.

\begin{figure}
\begin{center}
\includegraphics[width=.9\linewidth]{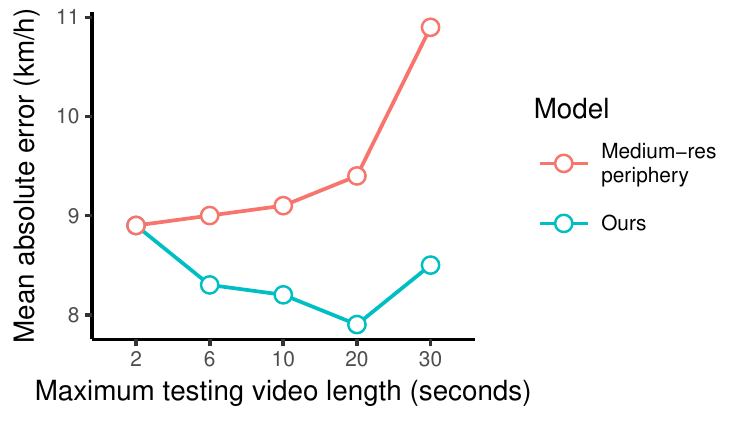}
\end{center}
   \caption{Testing errors of the medium-resolution periphery-only model and our model calculated using different lengths of testing videos. The two models have the same amount of FLOPs at inference time, but our model consistently showed greater driving accuracy than the competing model.}
\label{fig:mid_res_error}
\end{figure}

\subsection{Multi-resolution vs. Uni-resolution}
We further compared the performance of our periphery-fovea multi-resolution model with an uni-resolution periphery-only design, \ie, allocating all the resources to increase the resolution of the periphery vision without adding foveal vision. The number of floating-point operations (FLOPs) of our multi-resolution model for processing every video frame at inference is 3.4 billion. A medium-resolution periphery-only model that matches the same amount of FLOPs has a periphery input resolution size of $209\times371$ pixels. The structure of this model was the same as the periphery branch of our model except one change due to the enlarged input resolution. The periphery encoder of our model output feature maps of $3\time7$ pixels and then upsampled them to $9\times16$ pixels. The periphery encoder of the medium-resolution model output feature maps of $12\times22$ pixels and then downsampled them to $9\times16$ pixels. We tested this medium-resolution periphery-only (medium-res periphery) model against our periphery-fovea multi-resolution model. For a thorough analysis, we did the comparison for multiple rounds. In each round we cut the test videos into segments no longer than a certain length and tested the models using those segments. We tried segment lengths from two seconds up to 30 seconds (the longest single segment that we could process with our GPU memory). The prediction errors of the two models measured in MAE are summarized in Figure~\ref{fig:mid_res_error}. The prediction error of the medium-res periphery model kept increasing with increasing video length, while the prediction error of our model stayed more stable. Our model showed smaller prediction errors than the medium-res periphery model with all video lengths except with 2 seconds the two models showed the same error. Over all, the result suggested that the periphery-fovea multi-resolution design would achieve better driving accuracy than a uni-resolution periphery-only design given the same amount of computation.


\section{Conclusion}
We have proposed a new periphery-fovea multi-resolution driving model that combines global low-resolution visual input and local high-resolution visual input. We have shown that guiding the foveal vision module by predicted human gaze significantly improves driving accuracy with high efficiency. The performance gain is even more significant in pedestrian-involved critical situations than other average driving situations. Our approach has demonstrated a promising avenue to incorporate human attention into autonomous driving models to handle crucial situations and to enhance the interpretability of the model's decisions.

\normalsize
\bibliography{references}


\end{document}